\documentclass{article}

\usepackage[main, final]{neurips_2026}
\makeatletter
\renewcommand{\@notice}{}
\makeatother

\usepackage[utf8]{inputenc} 
\usepackage[T1]{fontenc}    
\usepackage{hyperref}       
\usepackage{url}            
\usepackage{booktabs}       
\usepackage{amsfonts}       
\usepackage{nicefrac}       
\usepackage{microtype}      
\usepackage{xcolor}         
\usepackage{amsmath}
\usepackage{graphicx}
\usepackage{subcaption}
\usepackage{algorithm}
\usepackage{algpseudocode}
\usepackage{xcolor}
\usepackage{wrapfig}
\usepackage{multirow}
\usepackage{booktabs}
\usepackage{hyperref}
\usepackage{amsthm}

\usepackage{subcaption}
\usepackage[table]{xcolor}

\usepackage{xcolor}
\usepackage{xcolor}

\definecolor{suffcolor}{HTML}{EAD4D1}
\definecolor{neccolor}{HTML}{D6E2EA}  

\newtheorem{theorem}{Theorem}[section]

\newtheorem{definition}[theorem]{Definition}

\title{Beyond Sufficiency: Time Series Explanation with Counterfactual Necessity}

\author{%
Hongnan Ma\textsuperscript{1},
Yiwei Shi\textsuperscript{2},
Mengyue Yang\textsuperscript{2},
Weiru Liu\textsuperscript{2}
\\[4pt]
\textsuperscript{1}School of Computer Science, University of Bristol
\\
\textsuperscript{2}School of Engineering Mathematics and Technology, University of Bristol
\\[4pt]
\texttt{hongnan.ma@bristol.ac.uk}
\\
\texttt{\{yiwei.shi, mengyue.yang, weiru.liu\}@bristol.ac.uk}
}

\begin{document}

\maketitle

\begin{abstract}
Faithful explanations of time-series classifiers should identify subsequences that are not only sufficient to preserve a black-box model's prediction, but also necessary for maintaining it. However, existing sufficiency-oriented methods can assign high importance to spurious subsequences that support the prediction without being essential to the model's decision. We introduce \textbf{TimePNS}, a necessity-aware framework for time-series explanation. Inspired by Pearl's counterfactual notion of necessity, TimePNS assesses whether a temporal factor is necessary by intervening on it and measuring whether the original prediction is disrupted. The framework adopts a two-stage design. Stage I learns an identifiable causal generative process together with a sufficiency-oriented explanation mask. Stage II performs counterfactual interventions on temporal factors to derive necessity signals, which supervise a temporal gate that refines the initial explanation by suppressing non-essential components and emphasizing counterfactually necessary ones. Experiments on synthetic and real-world time-series benchmarks show that TimePNS more accurately identifies decision-critical subsequences and consistently improves sufficiency-necessity trade-offs over strong baselines.
\end{abstract}



\section{Introduction}
Explaining predictions made by black-box models for multivariate time-series is a fundamental requirement in high-stakes applications such as healthcare, finance, and autonomous systems~\citep{morid2023time,kim2025comprehensive}. In these settings, practitioners must identify which variables and time steps are truly responsible for a model’s output in order to support timely intervention and build trust in automated decision systems~\citep{kusters2020conceptual}. Most existing explanation methods tackle this problem by applying masking and perturbation operations to the input sequence and measuring the resulting change in prediction to infer the importance score of different variables or temporal segments~\citep{crabbe2021explaining}. These methods aim to extract a subset of the input sequence that preserves the original prediction while remaining concise and interpretable.

However, when viewed through the lens of the probability of necessity and sufficiency (PNS)~\citep{pearl2009causality}, existing methods primarily operationalize \textit{sufficiency}: they seek subsequences that are sufficient to preserve the model’s original prediction when retained~\citep{queen2023encoding,liu2024timex++}. This limitation becomes particularly problematic in real-world time series, where different temporal segments or channels may be highly correlated, giving rise to spurious associations that make multiple subsequences appear informative for the model’s prediction~\citep{zhang2023towards}. In such cases, a sufficiency-based explanation method may assign high importance to correlated but non-essential subsequences, leading to explanations that do not faithfully reflect what the model actually relies on. When applied to new time series data, such fragile correlations may no longer persist, making these explanations unreliable under distribution shift~\citep{yang2023invariant, scholkopf2021toward}. In contrast, the complementary notion of \textit{necessity} provides an additional criterion for filtering out correlated but non-essential subsequences.

In Pearl’s framework, necessity is captured through a counterfactual `what-if' question: would the outcome fail to occur if the factor of interest were absent~\citep{pearl2009causality}? This counterfactual view sharpens the notion of time series explanation: a factor is not explanatory simply because it can preserve the black-box model’s original prediction, but because removing it changes that prediction. Fig.~\ref{motivation} illustrates this distinction with an ECG example for myocardial infarction detection (MI). The ST-elevation segment in the affected leads is the diagnostically decisive evidence, whereas the reciprocal ST-depression observed in the opposing leads is a frequent co-occurring pattern that is highly correlated with the same underlying condition and therefore also appears predictive to the model. Under a sufficiency-based criterion, both subsequences may be considered adequate explanations, since either alone can preserve the original prediction. A necessity-based analysis, however, reveals a clear asymmetry: in a model that has learned to rely primarily on ST-elevation, masking this segment flips the prediction, whereas masking the correlated reciprocal ST-depression leaves the prediction unchanged.


Motivated by this observation, we propose \textbf{TimePNS}, a framework for explanation of black-box time-series models that moves beyond the prevailing sufficiency-based paradigm.  However, distinguishing genuinely necessary temporal subsequences is particularly challenging in time series settings, where raw observations are often highly mixed and naive input-level counterfactual interventions may violate the underlying temporal  dependencies~\citep{li2024identification}. To address this challenge, TimePNS introduces a two-stage training procedure built upon a learned causal latent representation. In Stage~I, a mask-based explainer is trained to select compact subsequences that preserve the black-box model's original prediction, while a causal representation branch maps the input into a structured latent space and learns both instantaneous and lagged causal mechanisms among latent factors. The second stage refines the initial explanation from the perspective of necessity. Given the learned causal mechanisms, TimePNS estimates a soft probability of necessity score for each temporal factor through latent
counterfactual intervention. These scores are then used to supervise a temporal gate, which filters out factors with low estimated necessity and refines the explanation toward decision-critical subsequences. In this way, TimePNS encourages the final explanation to preserve the original prediction while being aligned with counterfactual necessity. The main contributions are summarized as follows: \textbf{1)} We show that existing time-series explanation methods may select predictive but non-essential subsequences, as they focus mainly on sufficiency rather than necessity.\textbf{ 2)} We propose \textbf{TimePNS}, a framework that estimates counterfactual necessity via structured interventions in a learned causal latent space. \textbf{3)} Experiments show that TimePNS improves critical-subsequence identification and suppresses spurious explanations compared with baselines.

\section{Related work}
\textbf{Time Series Explanation.}
Recent explanation methods for time-series models largely follow a perturbation-based paradigm~\citep{crabbe2021explaining}. One line of work directly perturbs the queried time series and feeds the perturbed inputs into the black-box model to estimate the importance of temporal regions or
time--feature pairs. Representative methods include
Dynamask~\citep{crabbe2021explaining} and Extremask~\citep{enguehard2023learning}, which learn instance-specific masks and evaluate their relevance through the induced changes in model predictions. Another line of work incorporates perturbations into
surrogate-based explanation models. TimeX~\citep{queen2023encoding} learns an interpretable surrogate via a self-supervised behavioral consistency objective, while TimeX++~\citep{liu2024timex++} further introduces an
information bottleneck objective and an explanation conditioner to generate in-distribution, label-preserving explanation instances. However, these
methods remain largely sufficiency-oriented: they identify regions that can preserve the original prediction, but do not determine whether those regions
are truly necessary for the model's decision.

\textbf{Causal Sufficient and Necessary.}
PNS was introduced in Pearl's structural causal model framework as one of the probabilities of causation, together with PN and PS~\citep{pearl2009causality}. Recently, necessity and sufficiency-based ideas have been adopted to identify invariant or robust predictive factors, including representation learning under distribution shift~\citep{yang2023invariant}, graph OOD generalization~\citep{chen2025unifying}. PNS has also been introduced into explanation and reasoning tasks. In
textual causal rationalization, rationale selection is formulated within a structural causal model, where conditional PNS is used to identify non-spurious rationales~\citep{zhang2023towards}. More recently,
PNS-based intervention analysis has been used to assess whether individual Chain-of-Thought reasoning steps are causally necessary and sufficient for the final answer~\citep{yu2025causal}. However, these applications typically operate on observed units, such as tokens or reasoning steps, which can be directly intervened upon. This observation-level formulation is less suitable for multivariate time-series explanation, where raw subsequences are often entangled mixtures of latent causal factors and input-space perturbations may violate instantaneous and lagged temporal dependencies. 

\section{Notation and Problem Setup}
We denote a multivariate time-series instance as $X \in \mathcal{X} = \mathbb{R}^{T \times D}$, where $T$ is the number of time steps and $D$ is the number of signals. The observation at time step $t$ is denoted by $x_t \in \mathbb{R}^{D}$, and a contiguous sub-sequence from time step $i$ to time step $j$ is denoted by $x_{i:j} \in \mathbb{R}^{(j-i+1)\times D}$. Let $F_{\phi}:\mathcal{X}\to\mathcal{Y}$ be a pretrained black-box predictor. We denote its predicted class by $\hat{y}(X) = \arg\max_{c} F_{\phi,c}(X)$ where $c$ presents class $c$. In post-hoc instance-level time-series explanation, the goal is to identify a subset of subsequences of $X$ that explain the prediction of $F_{\phi}$ on $X$.

\begin{definition}[Time-Series Mask Matrix]
\label{explanation}
Given an input time-series $X \in \mathbb{R}^{T \times D}$, we define a mask matrix $M \in \{0,1\}^{T \times D}$, where $M_{t,d}=1$ indicates that the corresponding subsequences are important and can be used for explanation. 



\end{definition}
For example, in Figure~\ref{motivation}, if the yellow and blue subsequences are assigned mask value 1, then these subsequences are selected as the explanation.



To obtain a meaningful explanation, we learn the explanation mask $M$ under a task-driven objective, such as the sufficiency and necessity objectives function introduced in Sec.~\ref{timepns}. 
Since directly optimizing over the discrete mask space is intractable, we also relax the problem into a continuous optimization via stochastic masking~\citep{queen2023encoding,liu2024timex++}. 
Specifically, an explanation extractor $g_\theta(\cdot)$ maps the input to Bernoulli parameters:
\begin{equation}
    \pi = g_\theta(X) \in [0,1]^{T\times D}, 
    \qquad 
    M_{t,d} \sim \mathrm{Bernoulli}(\pi_{t,d}).
    \label{maskgenerator}
\end{equation}


\section{Causal Sufficiency and Necessity for Time-Series Explanation}
\label{suffinecess}
We further define such an explanation inspired by Pearl's counterfactual notions of sufficient and necessary~\citep{pearl2009causality}. To formalize these concepts, we firstly introduce a pair of complementary interventions $\mathrm{do}(M)$  and $\mathrm{do}(\bar{M})$.

\begin{definition}[Mask-induced Intervention]
Given an input time series $X \in \mathbb{R}^{T\times D}$ and a mask matrix $M \in \{0,1\}^{T\times D}$ as in Definition~\ref{explanation}, we define a mask-induced intervention as replacing the subsequences with mask value $0$ by a baseline sample $b\in\mathbb{R}^{T\times D}$. 
\label{maskinduced}
\end{definition}

The baseline sample is drawn from a distribution $\mathbb{B}_{\mathcal X}$. In
practice, we approximate this distribution by a factorized Gaussian:
\begin{equation}
\mathbb{B}_{\mathcal X}
=
\prod_{t=1}^{T}\prod_{d=1}^{D}
\mathcal{N}\!\left(\mu_{t,d},\sigma_{t,d}^{2}\right),
\label{baseline}
\end{equation}
where $\mu_{t,d}$ and $\sigma_{t,d}^{2}$ are the empirical mean and variance at time step $t$ and feature dimension $d$, estimated from the training dataset.

We use $\mathrm{do}(M)$ to denote the time series obtained by applying the mask-induced intervention defined in Definition~\ref{maskinduced}. The corresponding prediction is denoted by $\hat y^{\mathrm{do}(M)}$. Conversely, we use $\mathrm{do}(\bar M)$ to denote the complementary intervention, where the subsequences assigned value 1 in $M$ are replaced with a baseline. The corresponding prediction is denoted by $\hat y^{\mathrm{do}(\bar M)}$.

\subsection{Definitions of Sufficiency and Necessity Explanation}
\begin{definition}[Probability of Sufficiency]
\label{def:sufficiency}
For an explanation mask $M$, the Probability of Sufficiency (PS) is defined as {$\mathrm{PS}(M)
=\mathbb{P}\!\left(
\hat{y}^{do(M)}=\hat{y}
\;\middle|\;
\hat{y}^{do(\bar{M})}\neq\hat{y}
\right)$},  whenever the conditioning event has nonzero probability. A high $\mathrm{PS}(M)$ indicates that the subset of subsequences assigned value 1 by $M$ alone are sufficient to preserve the original prediction.
\end{definition}

\begin{definition}[Probability of Necessity]
\label{def:necessity}
For an explanation mask $M$, the Probability of Necessity (PN) is defined as
{$ \mathrm{PN}(M)
= \mathbb{P}\!\left(
\hat{y}^{do(\bar{M})}\neq\hat{y}
\;\middle|\;
\hat{y}^{do(M)}=\hat{y}
\right)$}, whenever the conditioning event has nonzero probability.
A high $\mathrm{PN}(M)$ indicates that the subset of subsequences assigned value 1 by $M$ alone necessary in the sense that removing it changes the original prediction.
\end{definition}

\begin{definition}[Probability of Necessity and Sufficiency]
\label{def:pns}
For an explanation mask $M$, the Probability of Necessity and Sufficiency (PNS) is defined as
{$
\mathrm{PNS}(M)
=\mathbb{P}\!\left(
\hat{y}^{do(M)}=\hat{y},
\;\hat{y}^{do(\bar{M})}\neq\hat{y}
\right).$}
A high $\mathrm{PNS}(M)$ indicates that the subset of subsequences assigned value 1 by $M$ are both sufficient to preserve the original prediction and necessary as removing it changes the original prediction.
\end{definition}

\section{Counterfactual Reasoning for Necessary Explanations}
\label{PNintervention}
For sufficiency-based explanations, we follow the standard masking-based paradigm: an explanation extractor \(g_{\theta}\) learns a binary mask $M$ over the input time series such that the selected entries preserve the original prediction of the classifier $F_{\phi}$~\citep{queen2023encoding,liu2024timex++}. In contrast, necessary explanations are counterfactual: they evaluate whether the original prediction would change after removing the information selected by the explanation~\citep{pearl2009causality}. However, observed variables in time series are often entangled nonlinear mixtures of multiple latent factors \citep{hyvarinen2017nonlinear}. Directly intervening in the observation space simultaneously perturbs multiple latent factors and yield off-manifold counterfactuals. We therefore estimate PN in a latent causal space, where we model the latent generative process, formalize latent dependencies via inverse mechanisms, and propagate counterfactual interventions to compute PN scores.

\subsection{Latent Causal Generative Process}
\begin{wrapfigure}[16]{r}{0.35\columnwidth}
    \vspace{-8pt}
    \centering
    \includegraphics[
        width=\linewidth,
        trim=10pt 8pt 10pt 8pt,
        clip
    ]{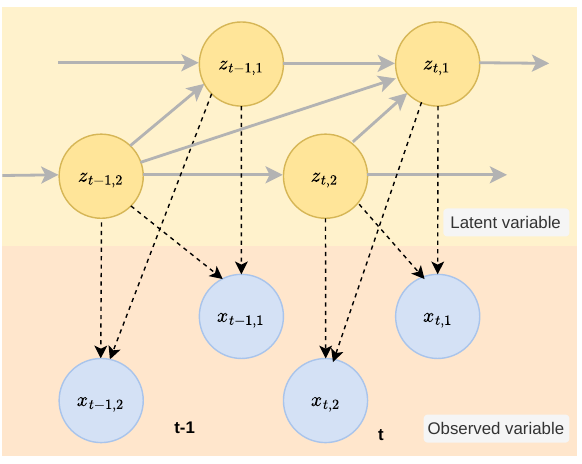}
    \vspace{-10pt}
    \caption{Latent variable generative process. The black dashed lines represent the latent causal process, while the gray lines indicate the mixture process.}
    \label{data_process}
    \vspace{-10pt}
\end{wrapfigure}

We model a multivariate time series as a nonlinear observation of an underlying latent causal process, following~\citep{li2024identification}. Let \(X=\{x_t\}_{t=1}^T\) denote an observed time series, where \(x_t\in\mathbb{R}^D\). Each observed variable is assumed to be generated from a latent state \(z_t=(z_{t,1},\ldots,z_{t,d})^\top\in\mathbb{R}^d\) through a nonlinear observation function \(h:\mathbb{R}^d\rightarrow\mathbb{R}^D\), such that \(x_t=h(z_t)\) for \(t=1,\ldots,T\).

The latent generative process is governed by a temporal structural causal model (SCM) with both lagged and instantaneous dependencies, as illustrated in Fig.~\ref{data_process}. For each
latent component \(z_{t,i}\), its lagged parents are drawn from previous latent states within the lag window, while its instantaneous parents come from other components at the same time step. Specifically,

\begin{equation}
    z_{t,i}
    =
    f_i\!\left(
        \mathrm{Pa}^{\mathrm{lag}}(z_{t,i}),
        \mathrm{Pa}^{\mathrm{inst}}(z_{t,i}),
        \epsilon_{t,i}
    \right),
    \qquad
    \epsilon_{t,i}\sim p_{\epsilon_i},
    \label{eq:scm}
\end{equation}
The exogenous noises \(\epsilon_{t,i}\) are mutually independent over time and latent dimensions.

\subsection{Learning an Inverse Mechanism for Latent Dynamics}
\label{sec:counterfactual_reasoning}
Rather than explicitly learning the unknown forward structural mechanisms \(f_i\) in Eq.~\eqref{eq:scm}, we adopt an inverse-mechanism formulation,
following~\citep{li2024identification}. For each latent component \(z_{t,i}\), we learn a neural network \(r_i\) that maps the current latent state and its lagged context to a recovered residual:
\begin{equation}
    \hat{\epsilon}_{t,i}
    =
    r_i\!\left(z_{t-L:t-1},\, z_t\right),
    \label{eq:residual_fn}
\end{equation}
where \(z_{t-L:t-1}\) denotes the latent history within the maximum lag \(L\), and \(\hat{\epsilon}_{t,i}\) is the residual recovered for the \(i\)-th latent component.

The inverse-mechanism formulation provides a local characterization of latent dependencies through the Jacobian of \(r_i\). Specifically, derivatives with respect to lagged latent variables, \(\frac{\partial r_i}{\partial z_{t-\ell,j}}\), indicate time-delayed dependencies, whereas off-diagonal derivatives with respect to contemporaneous latent variables, \(\frac{\partial r_i}{\partial z_{t,j}}\) for \(j\neq i\), characterize instantaneous dependencies~\citep{li2024identification}.


\subsection{Latent Counterfactual Propagation}
Given the factual latent sequence \(Z=\{z_t\}_{t=1}^{T}\) obtained by encoding \(X\) with \(\mathrm{enc}_{\eta}\), and the learned inverse mechanisms \(\{r_i\}_{i=1}^{d}\), we construct a counterfactual latent sequence \(Z^{\mathrm{cf}(\tau,k)}\) by intervening on a single latent
factor \(z_{\tau,k}\), which is then propagated through the learned latent dependency structure. Here, \(\tau\) denotes the intervention time, \(k\) denotes the intervened latent factor, and \(s\geq\tau\) denotes a time step reached during counterfactual propagation. More concretely, for each latent factor \(i\), we first compute its factual recovered residual at every time step:
$
    \hat{\epsilon}^{\mathrm{factual}}_{s,i}
    =
    r_i\!\left(z_{s-L:s-1},\, z_s\right),
     s=1,\dots,T .
$ We then intervene on \(z_{\tau,k}\) by setting it to a reference value \(\tilde z_{\tau,k}\) via \(\mathrm{do}(z_{\tau,k}=\tilde z_{\tau,k})\). Under the SCM counterfactual formulation, this intervention
replaces the structural equation of \(z_{\tau,k}\) with the assignment \(z_{\tau,k}=\tilde z_{\tau,k}\), while the recovered residuals of all non-intervened latent variables are held fixed at their factual values. Under the topological ordering of the instantaneous graph, an intervention on \(z_{\tau,k}\) can propagate only to variables ordered after \(k\) at the same time step and to variables at future time steps. Thus, states before
\(\tau\) and variables ordered before \(k\) at time \(\tau\) remain unchanged: $
    z^{\mathrm{cf}}_{s}=z_s \;\; (s<\tau), 
    z^{\mathrm{cf}}_{\tau,i}=z_{\tau,i} \;\; (i<k), 
    z^{\mathrm{cf}}_{\tau,k}=\tilde z_{\tau,k}.
$ Let $ \mathcal{D}_{\tau,k} = \{(\tau,i): i>k\}
    \cup
    \{(s,i): s>\tau,\;1\leq i\leq d\}
$ denote the set of potentially affected non-intervened variables. For these variables, we keep the recovered residuals fixed to their factual values:
\begin{equation}
    r_i\!\left(
        z^{\mathrm{cf}(\tau,k)}_{s-L:s-1},
        z^{\mathrm{cf}(\tau,k)}_{s}
    \right)
    =
    \hat{\epsilon}^{\mathrm{factual}}_{s,i},
    \qquad (s,i)\in\mathcal{D}_{\tau,k}.
    \label{eq:cf_implicit}
\end{equation}

To obtain a tractable local propagation rule, we linearize
Eq.~\eqref{eq:cf_implicit} around the factual latent sequence. Let \(\Delta z_{s,i}=z^{\mathrm{cf}(\tau,k)}_{s,i}-z_{s,i}\), and define
$J^{(\ell)}_{i,j}(s)=
\left.
\frac{\partial r_i}{\partial z_{s-\ell,j}}
\right|_{(z_{s-L:s-1},z_s)}
$
for lagged variables and $
J^{(0)}_{i,j}(s)=
\left.
\frac{\partial r_i}{\partial z_{s,j}}
\right|_{(z_{s-L:s-1},z_s)}
$ for instantaneous variables. Assuming \(J^{(0)}_{i,i}(s)\neq0\), the perturbation of each potentially affected variable is computed as
\begin{equation}
    \Delta z_{s,i}
    =
    -
    {(
        \displaystyle
        \sum_{\ell=1}^{L}\sum_{j=1}^{d}
        J^{(\ell)}_{i,j}(s)\Delta z_{s-\ell,j}
        +
        \sum_{j<i}
        J^{(0)}_{i,j}(s)\Delta z_{s,j}
    )}/{
        \displaystyle
        J^{(0)}_{i,i}(s)
    },
    \qquad (s,i)\in\mathcal{D}_{\tau,k}.
    \label{eq:cf_prop}
\end{equation}
The first term in the numerator captures propagation through lagged dependencies, while the second term captures propagation throug instantaneous dependencies under the topological ordering. The derivation from the linearized residual-preservation constraint is provided in Appendix~\ref{app:cf_propagation}.



\subsection{Counterfactual Necessity Scoring}
The counterfactual latent sequence is decoded back to the observation space: $X^{\mathrm{cf}(\tau,k)}
    =
    \left\{
    \mathrm{dec}_{\gamma}
    \!\left(
        z^{\mathrm{cf}(\tau,k)}_s
    \right)
    \right\}_{s=1}^{T}.
    \label{eq:cf_decode}$ The counterfactual sample \(X^{\mathrm{cf}(\tau,k)}\) is then passed to the
black-box model \(F\) for necessity scoring. For an input \(X\), We define the necessity score of latent factor
\(z_{\tau,k}\) as
\begin{equation}
    \Delta p_{\tau,k}
    =
    \left[
        F_{\phi,\hat y(X)}(X)
        -
        F_{\phi,\hat y(X)}
        \!\left(
            X^{\mathrm{cf}(\tau,k)}
        \right)
    \right]_+,
    \qquad
    [a]_+=\max(a,0).
    \label{eq:latent_pn_score}
\end{equation}


A larger \(\Delta p_{\tau,k}\) indicates that intervening on
\(z_{\tau,k}\) leads to a larger reduction in the factual-class confidence, suggesting that this latent factor is more necessary for maintaining the original prediction.

\section{TimePNS - PNS Estimation for Faithful Time Series Explanation}
\label{timepns}
Directly optimizing the joint PNS is computationally prohibitive. We therefore decouple the estimation into a two-stage procedure: Stage~I learns a sufficient explanation jointly with a structured latent causal model, and Stage~II uses counterfactual interventions on that latent causal model to derive a PN signal that
refines the explanation.


\subsection{Stage I: Latent Causal Mechanism Learning and Sufficiency-Oriented Explanation}
\label{sec:stage1}

\textbf{Latent Causal Mechanism Learning~\citep{li2024identification}.}
\label{sec:stage1_causal}
This branch trains the latent causal model introduced in Sec.~\ref{sec:counterfactual_reasoning}, which provides the structured intervention space used in Stage~II. Given an input time series \(X\), the encoder \(\mathrm{enc}_{\eta}\) parameterizes an approximate posterior \(q_{\eta}(Z\mid X)\) over the latent trajectory \(Z=(z_1,\dots,z_T)\). A latent trajectory sampled from this posterior is decoded to reconstruct the input, $
    \tilde X=\mathrm{dec}_{\gamma}(Z).
$ Beyond reconstruction, we regularize the latent dynamics with the inverse transition mechanisms \(\{r_i\}_{i=1}^{d}\) introduced in Eq.\ref{eq:residual_fn}. The Jacobian entries of these inverse mechanisms define local instantaneous and lagged dependency scores, which are later reused to propagate latent counterfactual interventions in Stage~II. The causal branch is optimized by
\begin{equation}
\mathcal{L}_{\mathrm{causal}}
=
\lambda_{\mathrm{rec}}\mathcal{L}_{\mathrm{rec}}
+
\lambda_{\mathrm{kld}}
\left(
\mathcal{L}_{\mathrm{kld}}^{\mathrm{normal}}
+
\mathcal{L}_{\mathrm{kld}}^{\mathrm{future}}
\right)
+
\lambda_{\mathrm{sp}}
\left(
\|J^{\mathrm{inst}}\|_{1}
+
\sum_{\ell=1}^{L}\|J^{\mathrm{lag}(\ell)}\|_{1}
\right),
\label{eq:causal_loss}
\end{equation}
Here, \(J^{\mathrm{inst}}\) and \(J^{\mathrm{lag}(\ell)}\) collect the instantaneous and \(\ell\)-lag Jacobian entries of the inverse mechanisms across time. The reconstruction term preserves information about the input, the KL terms regularize the recovered residuals under the temporal causal prior, and the \(L_1\) penalty encourages sparse dependency structures. Detailed definitions of the posterior, residual-based KL terms, temporal causal prior likelihood, and Jacobian penalties are provided in Appendix~\ref{app:latent_causal_learning}.

\textbf{Sufficiency-Oriented Explanation.}
\label{sec:stage1_explanation}
Stage~I trains the explanation branch using a sufficiency-oriented masking objective built on the mask formulation in Definition~\ref{explanation}. The explanation extractor \(g_{\theta}\) outputs Bernoulli parameters \(\pi=g_{\theta}(X)\in[0,1]^{T\times D}\), which define a factorized mask distribution
\(\mathbb{P}_{\pi}(M\mid X)=\prod_{t=1}^{T}\prod_{d=1}^{D}\mathrm{Bernoulli}(M_{t,d};\pi_{t,d})\).
A hard binary mask \(M\sim\mathbb{P}_{\pi}(\cdot\mid X)\) is sampled during the forward pass, and gradients are passed through the mask probabilities using a straight-through estimator (STE)~\citep{jang2016categorical}. To avoid the trivial all-one mask, we regularize the mask distribution toward a sparse Bernoulli prior \(\mathbb{Q}_{r}(M)=\prod_{t,d}\mathrm{Bernoulli}(M_{t,d};r)\):
\begin{equation}
\mathcal{L}_{\mathrm{mask}}
=
D_{\mathrm{KL}}\!\left(
\mathbb{P}_{\pi}(M\mid X)
\,\|\, 
\mathbb{Q}_{r}(M)
\right)
=
\sum_{t,d}
\left[
\pi_{t,d}\log\frac{\pi_{t,d}}{r}
+
(1-\pi_{t,d})\log\frac{1-\pi_{t,d}}{1-r}
\right],
\label{eq:mask_kl}
\end{equation}
where \(r\in(0,1)\) controls the target sparsity.

We further encourage temporal smoothness by applying a total-variation penalty to the mask probabilities,
\(\mathcal{L}_{\mathrm{con}}=\frac{1}{D(T-1)}\sum_{d=1}^{D}\sum_{t=1}^{T-1}|\pi_{t+1,d}-\pi_{t,d}|\). Predictive sufficiency is enforced by encouraging the masked input to preserve the frozen black-box predictive distribution,
\[
\mathcal{L}_{\mathrm{js}}
=
\mathbb{E}_{M\sim\mathbb{P}_{\pi}(\cdot\mid X),\,b\sim\mathbb{B}_{\mathcal X}}
\left[
D_{\mathrm{JS}}\!\left(
F_{\phi}(X)
\,\|\, 
F_{\phi}\!\left(X_{M}(b)\right)
\right)
\right],
\]
where \(X_M(b)=M\odot X+\bar{M}\odot b\) and \(\bar{M}=1-M\).

\textbf{Stage~I Training Objective.}
During Stage~I, the causal branch minimizes $\mathcal{L}_{\mathrm{causal}}$, while the explanation branch minimizes the sufficiency-oriented objective $\mathcal{L}_{\mathrm{suff}}=\lambda_{\mathrm{js}}\mathcal{L}_{\mathrm{js}}+\lambda_{\mathrm{mask}}\mathcal{L}_{\mathrm{mask}}+\lambda_{\mathrm{con}}\mathcal{L}_{\mathrm{con}}$. Here, $\mathcal{L}_{\mathrm{js}}$, $\mathcal{L}_{\mathrm{mask}}$, and $\mathcal{L}_{\mathrm{con}}$ correspond to prediction preservation, sparsity, and temporal smoothness, respectively, with $\lambda_{\mathrm{js}}$, $\lambda_{\mathrm{mask}}$, and $\lambda_{\mathrm{con}}$ controlling their relative weights.

\subsection{Stage~II: Necessity-Guided Explanation}
\textbf{Necessity Supervision Signal.}
After Stage~I pretraining, we use latent counterfactual interventions to construct a necessity supervision signal. Let \(\hat y=\arg\max_c F_{\phi,c}(X)\) denote the original black-box prediction. Necessity refinement is applied only when the Stage~I explanation preserves this prediction, indicated by \(I_{\mathrm{suff}}=\mathbf{1}[\arg\max_c F_{\phi,c}(X_M)=\hat y]\). For each intervention time \(\tau\in\mathcal{T}_{\mathrm{eval}}\) and latent factor \(k\), we convert the counterfactual probability drop \(\Delta p_{\tau,k}\) into a soft necessity target \(\widetilde{\mathrm{PN}}_{\tau,k}=I_{\mathrm{suff}}\Delta p_{\tau,k}\in[0,1]\). This target supervises the temporal gate by assigning higher preservation weights to time-feature factor whose intervention substantially reduces the factual-class confidence.

\textbf{Necessity-Guided Temporal Gate.} We introduce a lightweight gate $\rho_{\psi}$ over the causal latent trajectory. Given
\(Z^c\in\mathbb{R}^{T\times d}\), the gate outputs
\(G=\rho_{\psi}(Z^c)\in[0,1]^{T\times d}\) and produces the gated latent
trajectory \(Z^{\mathrm{gate}}=G\odot Z^c\), where \(G_{\tau,k}\) controls how
much the latent factor \(z^c_{\tau,k}\) is preserved.

The gate is trained with two objectives. First, for the sufficiency-valid mini-batch $\mathcal{B}_{\mathrm{suff}}=\{b:I_{\mathrm{suff},b}=1\}$, we use a
classification-anchor loss $\mathcal{L}_{\mathrm{clf}}^{\mathrm{gate}}
=
\frac{
\sum_{b\in\mathcal{B}_{\mathrm{suff}}}
\mathrm{CE}\!\left(
F_{\phi}(\mathrm{dec}_{\gamma}(Z^{\mathrm{gate}}_b)),
\hat y(X_b)
\right)
}{
\max(|\mathcal{B}_{\mathrm{suff}}|,1)
}.
\label{eq:gate_clf}
$

This loss encourages the decoded gated trajectory to preserve the original black-box prediction. Second, we align the temporal pattern of the gate with the soft necessity target:
$
\mathcal{L}_{\mathrm{align}}
=
\frac{
\sum_{b=1}^{B}\sum_{k=1}^{d}
I_{\mathrm{suff},b}
\left[
1-
\cos_{\mathcal{T}}\!\left(
G_{b,\mathcal{T}_{\mathrm{eval}},k},
\widetilde{\mathrm{PN}}_{b,\mathcal{T}_{\mathrm{eval}},k}
\right)
\right]
}{
d\cdot \max\!\left(\sum_{b=1}^{B} I_{\mathrm{suff},b},1\right)
}.
\label{eq:gate_align}
$
Here, \(\cos_{\mathcal{T}}\) denotes the centered cosine similarity computed over
the evaluated time steps \(\mathcal{T}_{\mathrm{eval}}\). Specifically, for two
temporal vectors \(u,v\in\mathbb{R}^{|\mathcal{T}_{\mathrm{eval}}|}\), it is
defined as  $
\cos_{\mathcal{T}}(u,v)
=
\frac{
(u-\bar u)^\top (v-\bar v)
}{
\|u-\bar u\|_2 \|v-\bar v\|_2 + \epsilon
},
$ where \(\bar u\) and \(\bar v\) are the temporal means of \(u\) and \(v\), and
\(\epsilon\) is a small constant for numerical stability. Finally, a linear projection of \(Z^{\mathrm{gate}}\) is injected into the representation of \(g_{\theta}\), allowing necessity-guided latent information to refine the final mask \(M\).

\textbf{Stage~II Training Objective.}
The PN-guided gate is optimized with $\mathcal{L}_{\mathrm{necessity}}=\mathcal{L}_{\mathrm{clf}}^{\mathrm{gate}}+\lambda_{\mathrm{cos}}\mathcal{L}_{\mathrm{align}}$, where $\mathcal{L}_{\mathrm{clf}}^{\mathrm{gate}}$ preserves the prediction, $\mathcal{L}_{\mathrm{align}}$ aligns the gate with necessity signals, and $\lambda_{\mathrm{cos}}$ controls the alignment strength.

\section{Experiment}
Experiments are structured around two core research questions:

\textbf{RQ1:} Does TimePNS produce sufficient and necessary explanations? (Sec. \ref{timepns_quality})    \\
\textbf{RQ2:} Does TimePNS refine the explanations by removing predictive but unnecessary part?(Sec.\ref{refine_explanation})



\subsection{Experiment setup}
We evaluate the quality of our explanations on three synthetic datasets and three real-world datasets. For each metric, \textbf{bold} denotes the best result and \underline{underline} denotes the second best. All results are reported as mean $\pm$ std over 5-fold cross-validation. All experiments in the main body use Vanilla Transformer~\citep{vaswani2017attention} with time-based positional encoding as the black-box classifier, with hyperparameter tuned to ensure strong predictive performance. 

\textbf{Dataset.} 
We evaluate the performance of our explainer on synthetic datasets with known ground-truth saliency scores from~\citep{queen2023encoding}: FreqShapes, SeqComb-MV, and Low-VAR. These datasets are carefully designed to cover diverse temporal dynamics in multivariate settings, where correctly identifying salient features at different timesteps is non-trivial. For real-world evaluation, we use three datasets from the UCR Archive~\citep{dau2019ucr}: Epilepsy, ERing, and ArticularyWordRecognition (WordRec), which span diverse temporal classification domains.


\textbf{Baselines.}
We compare our method with several baselines, including perturbation-based methods, namely TimeX~\citep{queen2023encoding}, TimeX++~\citep{liu2024timex++}, Dynamask~\citep{crabbe2021explaining}, as well as gradient-based methods, including CORTX~\citep{chuang2023cortx}, WinIT~\citep{leung2021temporal} and SGT+GRAD~\citep{ismail2021improving}, 

\textbf{Evaluation Metrics.}
We evaluate explanations under two settings. For synthetic datasets, we compare the generated saliency maps with the ground-truth predictive regions specified by the data-generating process. These regions serve as sufficient and necessary explanatory signals. Following \citet{crabbe2021explaining}, we report AUP, AUR, and AUPRC, where higher values indicate better agreement with the ground truth. For real-world datasets without ground-truth saliency, we use occlusion-based evaluation following~\citep{queen2023encoding}. Sufficiency is evaluated by retaining the top-\(p\) salient features and measuring prediction preservation, where higher AUROC and AUPRC is better. Necessity is evaluated by removing the top-\(p\) salient features and measuring the degradation of the remaining input, where lower AUROC and AUPRC is better. Detailed dataset descriptions, baseline implementations, and full metric definitions are provided in Appendix~\ref{app:baseline}.

\subsection{Experiment Result}
\subsubsection{TimePNS produce sufficient and necessary explanation}
\label{timepns_quality}
\begin{figure}[h]
    \centering
    \includegraphics[width=0.95\linewidth]{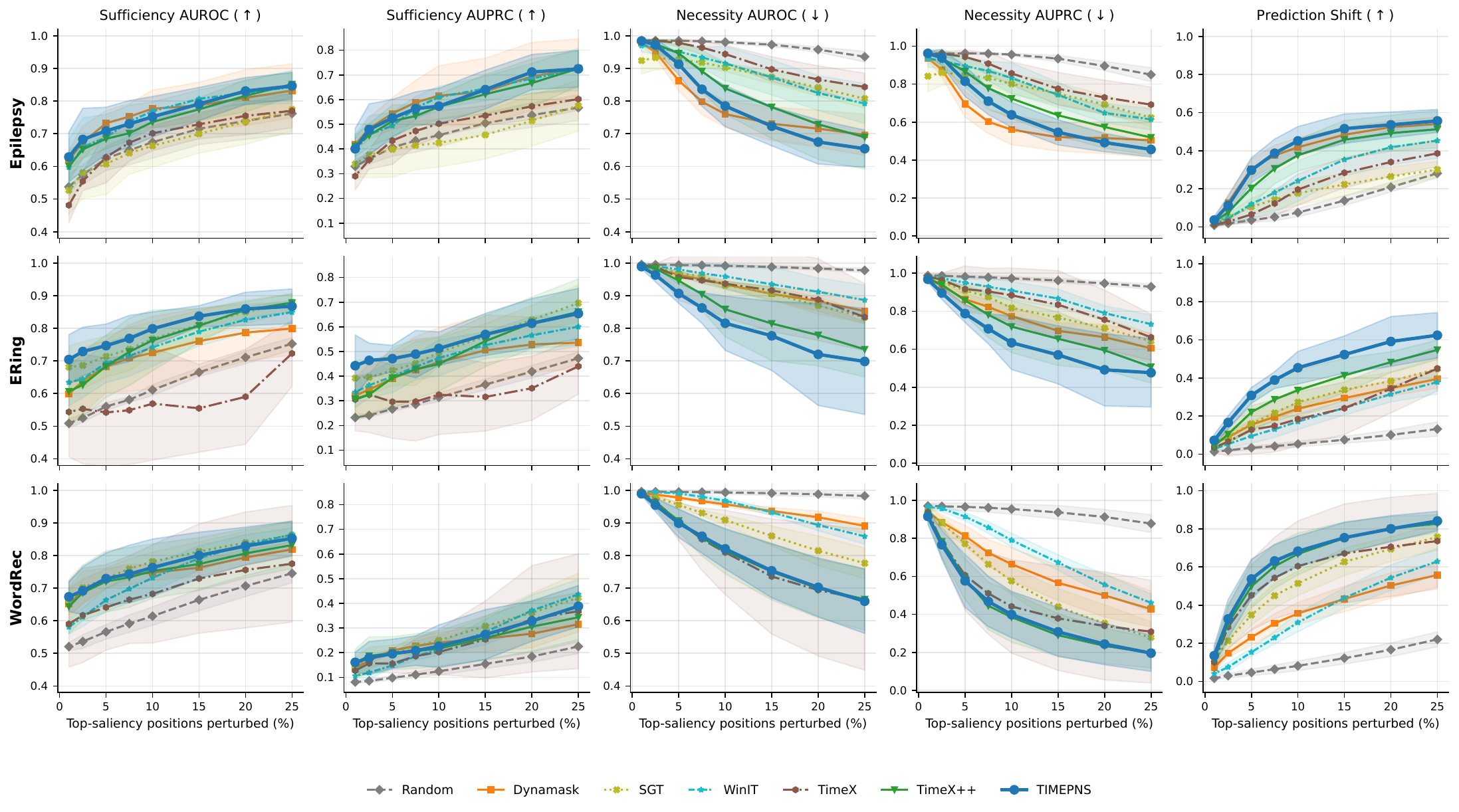}
    \caption{
Real-world occlusion evaluation on Epilepsy, ERing, and WordRec.
}
    \label{realworldresult}
\end{figure}

Table~\ref{three_datasets} reports explanation quality on synthetic datasets with ground-truth saliency. TimePNS achieves the best AUPRC on all three datasets and the best AUP on SeqComb-MV and FreqShape, showing that the proposed necessity-aware refinement improves the ranking and localization of salient regions. Compared with traditional methods, Dynamsk, WINIT, CORTX, and SGT show competitive results on individual metrics, such as SGT's AUR on SeqCombMV and CORTX's AUPRC on FreqShape, but their performance is less consistent across datasets and metrics. Among stronger recent baselines, TimeX and TimeX++ achieve the best AUR or AUP on LowVER, suggesting that sufficiency-oriented methods can be effective when the synthetic signal is clean and explicit. Nevertheless, TimePNS delivers the most consistent AUPRC gains, which is important for identifying compact and precise explanatory subsequences. For real-world datasets, where ground-truth saliency annotations are unavailable, we evaluate explanations using occlusion-based sufficiency and necessity tests, supplemented by Prediction Shift, measured as the total variation distance between the original and perturbed prediction distributions. As shown in Figure~\ref{realworldresult}, TimePNS achieves the best sufficiency performance across all datasets, indicating that its top-ranked positions preserve the black-box prediction most effectively. Under the necessity test, where lower AUROC and AUPRC indicate a larger performance drop after removing the selected positions, TimePNS induces the strongest degradation across datasets. It also yields the largest Prediction Shift, suggesting that perturbing its selected positions produces the most substantial change in the model output. These results demonstrate that TimePNS identifies positions that are both sufficient for preserving predictions and necessary for maintaining the model's original behavior.

\newcommand{\res}[2]{#1$_{\scriptscriptstyle \pm #2}$}
\newcommand{\best}[2]{\textbf{#1$_{\scriptscriptstyle \pm #2}$}}
\newcommand{\second}[2]{\underline{#1$_{\scriptscriptstyle \pm #2}$}}

\begin{table*}[h]
\centering
\footnotesize
\setlength{\tabcolsep}{3pt}
\renewcommand{\arraystretch}{1.08}
\resizebox{1.0\textwidth}{!}{
\begin{tabular}{l ccc ccc ccc}
\toprule
\rowcolor{gray!15}
& \multicolumn{3}{c}{SeqComb-MV} 
& \multicolumn{3}{c}{LowV-ER} 
& \multicolumn{3}{c}{FreqShape} \\
\cmidrule(lr){2-4}
\cmidrule(lr){5-7}
\cmidrule(lr){8-10}
Method 
& AUPRC ($\uparrow$) & AUP ($\uparrow$) & AUR ($\uparrow$)
& AUPRC ($\uparrow$) & AUP ($\uparrow$) & AUR ($\uparrow$)
& AUPRC ($\uparrow$) & AUP ($\uparrow$) & AUR ($\uparrow$) \\
\midrule

Dynamsk
& \res{0.3136}{0.0019} & \res{0.5481}{0.0053} & \res{0.1953}{0.0025}
& \res{0.1391}{0.0012} & \res{0.1640}{0.0028} & \res{0.2106}{0.0018}
& \res{0.2201}{0.0013} & \res{0.2952}{0.0037} & \res{0.5037}{0.0015} \\

WINIT
& \res{0.2809}{0.0018} & \res{0.7594}{0.0024} & \res{0.2077}{0.0021}
& \res{0.1667}{0.0015} & \res{0.1140}{0.0022} & \res{0.3842}{0.0017}
& \res{0.5071}{0.0021} & \res{0.5546}{0.0026} & \res{0.4557}{0.0016} \\

CORTX
& \res{0.3629}{0.0021} & \res{0.5625}{0.0006} & \res{0.3457}{0.0017}
& \res{0.4983}{0.0014} & \res{0.3281}{0.0027} & \res{0.4711}{0.0013}
& \res{0.6978}{0.0156} & \res{0.4938}{0.0004} & \res{0.3261}{0.0012} \\

SGT
& \res{0.4893}{0.0005} & \res{0.4970}{0.0005} & \best{0.4289}{0.0018}
& \res{0.3449}{0.0010} & \res{0.2133}{0.0029} & \res{0.3528}{0.0015}
& \res{0.5312}{0.0019} & \res{0.4138}{0.0011} & \res{0.3931}{0.0015} \\

TIMEX
& \second{0.6878}{0.0021} & \res{0.8326}{0.0008} & \second{0.3872}{0.0015}
& \res{0.8673}{0.0033} & \res{0.5451}{0.0028} & \best{0.9004}{0.0024}
& \second{0.8324}{0.0034} & \res{0.7219}{0.0031} & \res{0.6381}{0.0022} \\

TIMEX++
& \res{0.6663}{0.0021} & \second{0.8774}{0.0006} & \res{0.3329}{0.0012}
& \second{0.8957}{0.0020} & \best{0.8137}{0.0020} & \second{0.8313}{0.0020}
& \res{0.8286}{0.0022} & \second{0.7431}{0.0021} & \best{0.6426}{0.0021} \\

TimePNS
& \best{0.7460}{0.0015} & \best{0.9189}{0.0003} & \res{0.3156}{0.0009}
& \best{0.9188}{0.0029} & \second{0.7579}{0.0030} & \res{0.8034}{0.0027}
& \best{0.8574}{0.0026} & \best{0.7603}{0.0020} & \second{0.6396}{0.0021} \\

\bottomrule
\end{tabular}
}
\caption{Performance comparison on synthetic datasets.}
\label{three_datasets}
\end{table*}

\subsubsection{TimePNS refine the explanation}
\label{refine_explanation}
\begin{figure}[t]
    \centering
    \makebox[\textwidth][c]{%
    \begin{subfigure}[t]{0.45\textwidth}
        \vspace{0pt}
        \centering
        \includegraphics[width=\linewidth,height=3.8cm,keepaspectratio]{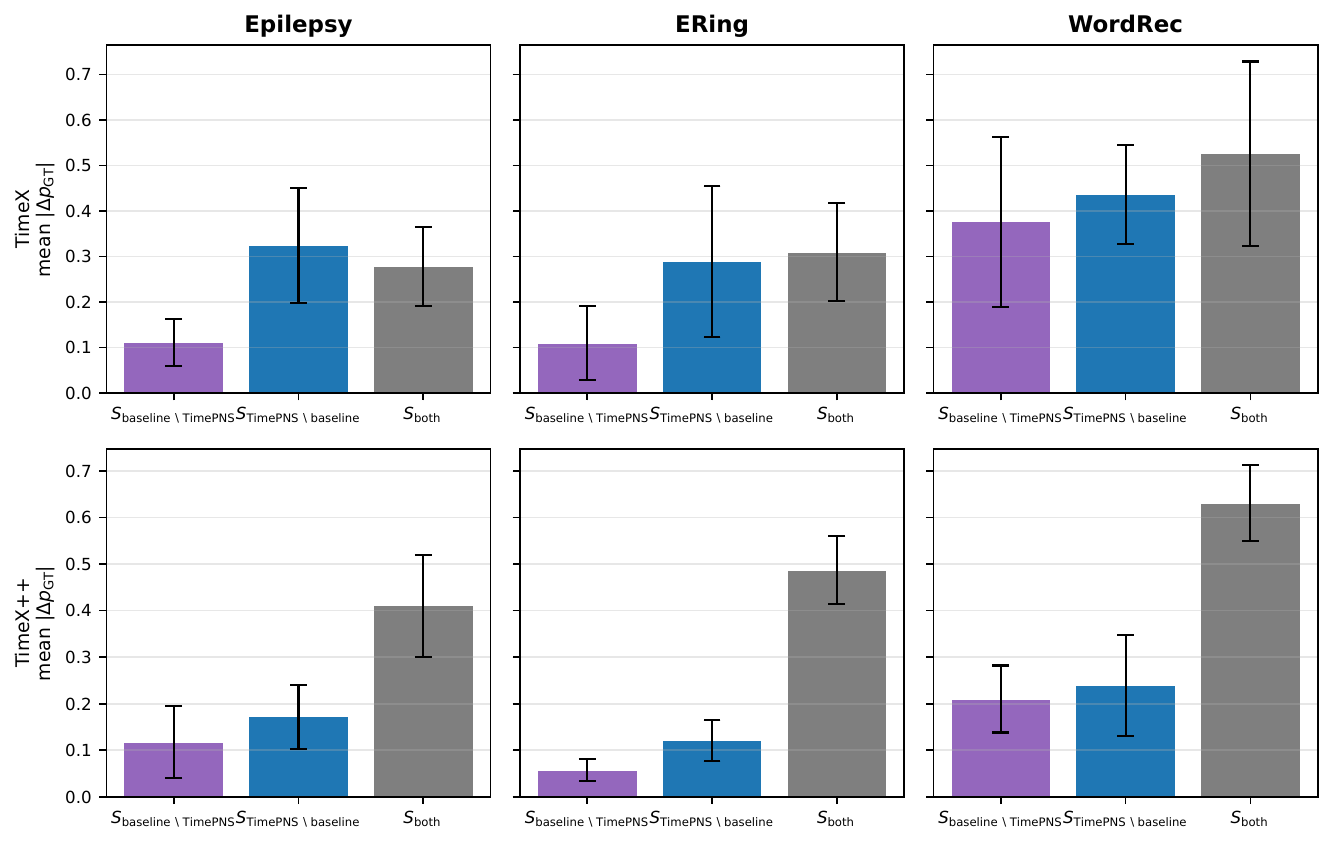}
        \caption{Disagreement-set ablation.}
        \label{fig:setdiff_combined}
    \end{subfigure}
    \hfill
    \begin{subfigure}[t]{0.45\textwidth}
        \vspace{0pt}
        \centering
        \includegraphics[width=\linewidth,height=3.8cm,keepaspectratio]{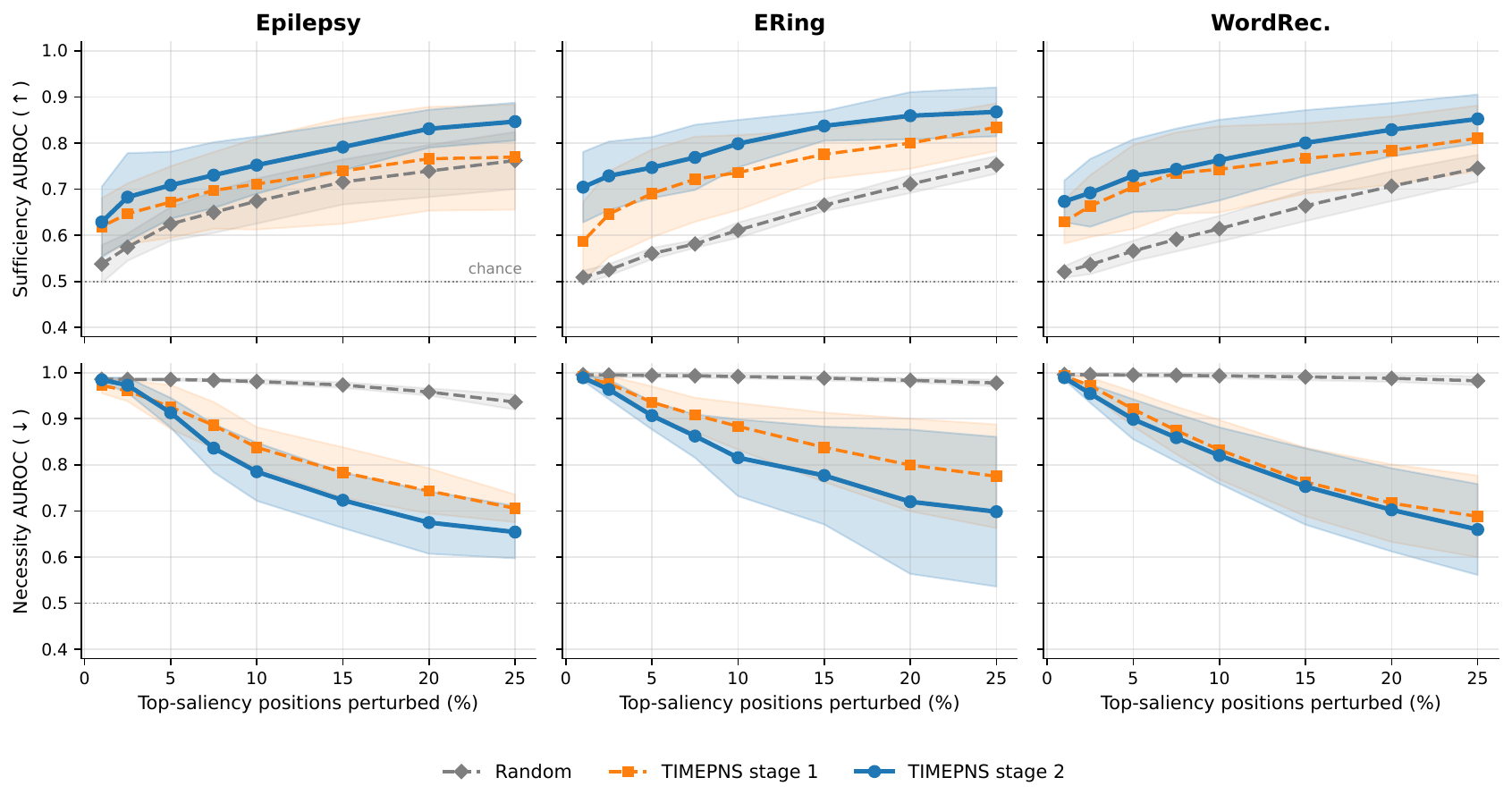}
        \caption{Stage I vs. Stage II.}
        \label{stage1_vs_stage2}
    \end{subfigure}
    }
    \caption{Ablation experiment.}
    \label{fig:additional_analysis}
\end{figure}

\paragraph{Disagreement-Set Ablation.}
To further assess whether TimePNS identifies features that are necessary for the black-box predictor, we conduct a disagreement-set ablation analysis. For each sample, we first select the top 25\% salient positions identified by the baseline (TimeX or TimeX++) and TimePNS, denoted as $\mathcal{S}_{\mathrm{baseline}}$ and $\mathcal{S}_{\mathrm{TimePNS}}$, respectively. We then construct three mutually disjoint subsets: positions selected only by the baseline, $\mathcal{S}_{\mathrm{baseline}}\setminus\mathcal{S}_{\mathrm{TimePNS}}$; positions selected only by TimePNS, $\mathcal{S}_{\mathrm{TimePNS}}\setminus\mathcal{S}_{\mathrm{baseline}}$; and positions selected by both methods, $\mathcal{S}_{\mathrm{both}} = \mathcal{S}_{\mathrm{baseline}}\cap\mathcal{S}_{\mathrm{TimePNS}}$.  We ablate each set by replacing its values with Gaussian samples estimated from the training data, and measure the resulting change in the ground-truth class probability,
$|\Delta p_{\mathrm{GT}}|$. As shown in Figure~\ref{fig:setdiff_combined}, TimePNS-exclusive positions induce larger prediction changes than baseline-exclusive positions in most settings, with the confident-sample win rate favoring TimePNS in five out of six comparisons. While the shared set often captures a common core of highly predictive features, the disagreement regions reveal the main difference: positions uniquely selected by TimePNS are more likely to be necessary for preserving the model's prediction than those uniquely selected by TimeX or TimeX++.

\textbf{Stage II Enhances Necessity-Aware Explanations} To isolate the effect of the second training stage, we compare the Stage~1 with the fully trained TimePNS using the same sufficiency and necessity protocol. As shown in Figure~\ref{stage1_vs_stage2}, the second stage consistently improves sufficiency across all real world dataset, indicating that the selected positions better preserve the black-box prediction. It also yields lower necessity AUROC, meaning that removing the selected positions causes a larger performance drop. This stronger improvement on the necessity side suggests that the necessary objective encourages saliency to concentrate on positions that the predictor relies on, rather than merely selecting sufficient or correlated regions. These results validate the role of the second stage as a necessity-aware refinement over the Stage~I.

\subsection{Case Study}
Figure~\ref{visulization} in Appendix~$\ref{vis}$ visualises a representative Epilepsy sample to illustrate the discrepancy between sufficiency and necessity. Using top-$50\%$ masks, all TimeX-family methods achieve high sufficiency scores, indicating that their selected timesteps can preserve the original prediction. However, their necessity behavior differs substantially: removing TimeX-selected timesteps causes almost no confidence drop ($0.001$), whereas removing TimeX++ and TimePNS masks leads to larger drops of $0.514$ and $0.740$, respectively. This shows that a sufficient mask can still be redundant, while TimePNS better identifies timesteps that are necessary for maintaining the model's prediction.

\section{Conclusion}
We introduced TimePNS, a necessity-aware framework for post-hoc explanation of time-series classifiers. Unlike sufficiency-oriented methods that mainly seek subsequences capable of preserving the black-box prediction, it further incorporates counterfactual necessity by refining a sufficiency-preserving mask with necessity signals derived from latent counterfactual interventions.  Experiments on synthetic and real-world benchmarks show that TimePNS improves the identification of decision-critical subsequences and achieves stronger sufficiency--necessity behavior than competitive baselines. Ablation analyses further validate the role of necessity-guided refinement in filtering out predictive but redundant regions. A limitation is the computational cost of repeated counterfactual interventions, which we leave to future work together with extensions to broader temporal architectures.

\newpage

\bibliographystyle{plainnat}
\bibliography{reference}

\clearpage
\appendix


\end{document}